%% file: root.tex
\documentclass[letterpaper, 10 pt, conference]{ieeeconf}  

\IEEEoverridecommandlockouts                              

\usepackage{amssymb}  
\usepackage{amsmath} 
\usepackage{threeparttable}
\usepackage{booktabs}
\usepackage{graphicx}
\usepackage{multirow}
\usepackage{makecell}
\usepackage{array}
\usepackage{cite}
\usepackage{booktabs}
\usepackage{hyperref} 
\usepackage{multirow}
\usepackage{colortbl}
\usepackage{xcolor}
\usepackage{dblfloatfix}
\usepackage{xspace}
\usepackage{array}
\usepackage{amsmath} 
\usepackage{amssymb} 
\usepackage{bm}       
\usepackage{graphicx}
\usepackage{placeins} 
\usepackage{threeparttable}
\title{\LARGE \bf
mmDEAR: mmWave Point Cloud Density Enhancement for Accurate Human Body Reconstruction
}

\author{Jiarui Yang$^{*}$ 
\quad
Songpengcheng Xia$^{*}$
\quad
Zengyuan Lai
\quad
Lan Sun
\quad
Qi Wu
\quad
Wenxian Yu
\quad
Ling Pei $^{\dagger}$
\thanks{* indicates these authors contributed equally to this work.}
\thanks{$\dagger$ Indicates the corresponding authors: Ling Pei (ling.pei@sjtu.edu.cn)}
\thanks{Jiarui Yang, Songpengcheng Xia, Zengyuan Lai, Lan Sun, Qi Wu, Wenxian Yu, and Ling Pei are with School of Electronic Information and Electrical Engineering, Shanghai Jiao Tong University, China. The work was supported by the National Natural Science Foundation of China under Grant 62273229.}}

\begin{document}

\maketitle
\thispagestyle{empty}
\pagestyle{empty}

\begin{abstract}
Millimeter-wave (mmWave) radar offers robust sensing capabilities in diverse environments, making it a highly promising solution for human body reconstruction due to its privacy-friendly and non-intrusive nature.  However, the significant sparsity of mmWave point clouds limits the estimation accuracy. To overcome this challenge, we propose a two-stage deep learning framework that enhances mmWave point clouds and improves human body reconstruction accuracy. Our method includes a mmWave point cloud enhancement module that densifies the raw data by leveraging temporal features and a multi-stage completion network, followed by a 2D-3D fusion module that extracts both 2D and 3D motion features to refine SMPL parameters. The mmWave point cloud enhancement module learns the detailed shape and posture information from 2D human masks in single-view images. However, image-based supervision is involved only during the training phase,  and the inference relies solely on sparse point clouds to maintain privacy. Experiments on multiple datasets demonstrate that our approach outperforms state-of-the-art methods, with the enhanced point clouds further improving performance when integrated into existing models.

\end{abstract}

\section{INTRODUCTION}
Human body reconstruction is a crucial research problem in the fields of computer vision and robotics \cite{chi2023pose, chen2023immfusion, zimmermann20183d, zhang2024dynamic}, with applications across various domains such as virtual/augmented reality \cite{zheng2023realistic,xia2024envposer}, human-robot interaction \cite{xia2024timestamp, 10161112}, humanoid robots \cite{zimmermann20183d, li2024fld} and health monitoring \cite{zhang2024dynamic, lai2024smart}. Current mainstream human body reconstruction methods can be categorized into three primary approaches: vision-based \cite{shen2023learning}, wearable-based \cite{zhang2024dynamic}, and wireless-based methods \cite{chen2023immfusion}.
While vision-based and wearable device-based approaches have shown impressive performance, they still face challenges in occlusion, privacy, and high intrusiveness \cite{hinojosa2021learning, chen2022mmbody, chen2024towards}.
\par
\begin{figure}[t]
    \centering
    \includegraphics[width=0.48\textwidth]{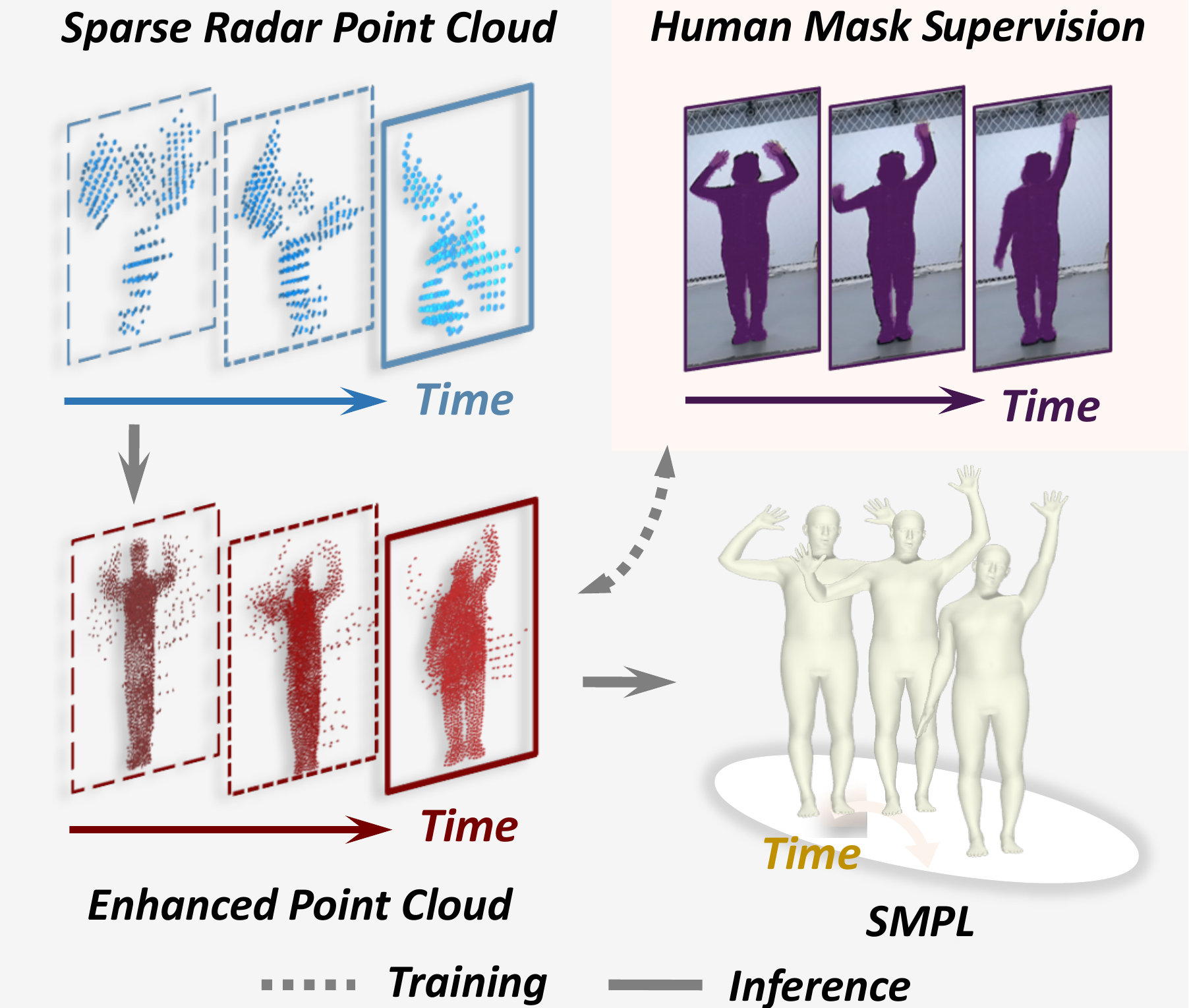} 
    \caption{
      Our proposed mmDEAR method first enhances the sparse point cloud to increase its density and capture detailed body posture and shape information, improving the accuracy of the final body reconstruction. The enhancement module is guided by a 2D human mask extracted from a single-view image during training, ensuring that the inference process is both image-free and privacy-preserving.
    }
    \label{fig:1}
    \vspace{-8mm}
    
\end{figure}

As a wireless device, mmWave radar offers privacy-preserving and non-intrusive solutions, mitigating many challenges faced by other sensing technologies. With the decreasing cost of mmWave radar and its advantages in detection range and angular resolution, it has become more popular in various fields, such as autonomous driving \cite{luan2024diffusion, huang2024multi} and smart homes \cite{zeller2024radar}. In human-centric tasks like tracking, action recognition, and gesture recognition, mmWave radar has demonstrated impressive performance \cite{palipana2021pantomime, singh2019radhar}. Despite the success of many studies utilizing mmWave point clouds for human body reconstruction \cite{chen2023immfusion, chen2022mmbody, yang2024mmbat, chen2024towards}, the sparse nature of these point clouds remains a significant challenge. A promising solution is to enhance the sparse mmWave point cloud by learning detailed human-specific information from easily obtained RGB images, and our strategy is shown in Fig. \ref{fig:1}.

Enhancing sparse point clouds generated by mmWave radar presents considerable technical challenges. Existing research on point cloud enhancement typically focuses on sensors such as LiDAR and RGB-D cameras, which aims to complete sparse and incomplete point clouds to reconstruct the full structure and surface of objects. However, there is limited research specifically addressing point cloud enhancement for mmWave radar, particularly in the context of human body reconstruction. We identify two key challenges in this task: First, mmWave point clouds exhibit inter-frame heterogeneity, in contrast to the more consistent data from LiDAR or RGB-D sensors. Second, conventional point cloud completion methods are primarily designed for static objects or autonomous driving scenes \cite{luan2024diffusion, xiang2022snowflake}, whereas our focus is on dynamic human bodies, which necessitates accounting for temporal changes and motion. To address these issues, we propose a multi-stage mmWave point cloud enhancement method that leverages 2D human mask information from single-view images as supervision during the training phase. By aggregating temporal features and employing a multi-stage estimation-refinement network, our approach generates enhanced mmWave point clouds specifically tailored for human body reconstruction.

Effectively extracting feature information from both raw and enhanced mmWave point clouds poses another significant challenge for achieving robust and accurate human body reconstruction. 
The raw mmWave point cloud provides more accurate and reliable 3D spatial information about the human body, while the enhanced point cloud offers more intuitive shape and structural details. Therefore, it is essential to leverage the motion features of both raw and enhanced point clouds. To address this challenge, we propose a multi-stage 2D-3D aware human body reconstruction module, which extracts 2D and 3D motion features from the combined point clouds and regresses the 2D and 3D joint positions separately. By fusing these extracted 2D and 3D features along with global features, we further refine the regression to obtain the final results.

In summary, the key contributions are as follows:

\begin{itemize}
\item We introduce an innovative two-stage deep-learning framework, named mmDEAR, designed for robust and accurate human body reconstruction using mmWave radar. This framework comprises a mmWave point cloud enhancement module and a 2D-3D aware human body reconstruction module.

\item Our proposed mmWave point cloud enhancement module incorporates a temporal feature extraction network to handle the inter-frame inconsistencies of mmWave point clouds, alongside a cascaded point cloud completion network that progressively densifies the sparse raw point clouds. Notably, this module uses the 2D human masks from single-view images for supervision during training but relies solely on sparse mmWave point clouds during inference, ensuring privacy and practical applicability.

\item Leveraging both enhanced and raw mmWave point clouds, we designed a 2D-3D aware human body reconstruction module that fully utilizes the dynamic 3D spatial information of the human body from the raw point cloud, and the detailed structural and shape information from the enhanced point cloud, resulting in improved body reconstruction accuracy.

\item Extensive experimental results on multiple mmWave datasets demonstrate that our proposed method outperforms the existing state-of-the-art (SOTA) methods. Furthermore, the enhanced point cloud produced by our mmWave point cloud enhancement module can be seamlessly integrated into other SOTA methods, resulting in further performance improvements.

\end{itemize}




















\section{Related Works}

\subsection{Human Body Reconstruction with mmWave Radar}
Due to its superior privacy and robustness in challenging environments, mmWave radar is becoming a popular choice for human body reconstruction tasks. Sengupta et al. \cite{sengupta2020mm} leverage two commercial mmWave radars from different angles to generate point clouds and estimate 3D skeletal joints by employing a simple 3D Convolutional Neural Network (CNN). mmMesh \cite{mmmesh} was the first method to regress the body parameters of the Skinned Multi-Person Linear (SMPL) model \cite{bogo2016SMPL} using mmWave radar data. In \cite{chen2022mmbody}, the author first adopted P4Transformer (P4Trans) \cite{fan2021p4trans} as the baseline to evaluate the performance of their dataset and obtained competitive results. In \cite{yang2024mmbat}, the researchers enhance the robustness of point cloud extraction from noisy data by predicting translation, which improves performance in challenging environments. Chen et al. \cite{chen2022mmbody} pioneered in combining mmWave radar with RGB camera, leading to significant improvements in human body reconstruction under challenging visual conditions such as low light, rain, and fog. Despite these advances, the inherent sparsity and inconsistency of mmWave point clouds continue to pose significant challenges for accurate human body reconstruction. 

\subsection{Point Cloud Enhancement}
Some recent works have successfully enhanced the mmWave radar point clouds during pre-processing and post-processing. Zhang et al. \cite{zhangmmeye} proposed a method to improve traditional Fast Fourier Transform (FFT) signal processing, while a learning-based approach \cite{brodeski2019deepradar} is adopted to avoid filtering out the real points. In \cite{luan2024diffusion}, the researchers enhanced mmWave radar data using supervision from LiDAR bird's-eye view (BEV) images. However, these approaches primarily target large-scale scenarios like autonomous driving and are not designed to capture the detailed pose and shape information essential for human body reconstruction. 
LiDAR point cloud completion methods \cite{xiang2022snowflake, cascaded} aim to recover accurate shapes from partial point clouds, similar to our goal, but they focus on static objects. Therefore, incorporating temporal information is crucial for enhancing dynamic human poses.

\begin{figure*}[t]
    \centering
    \includegraphics[width=\textwidth]{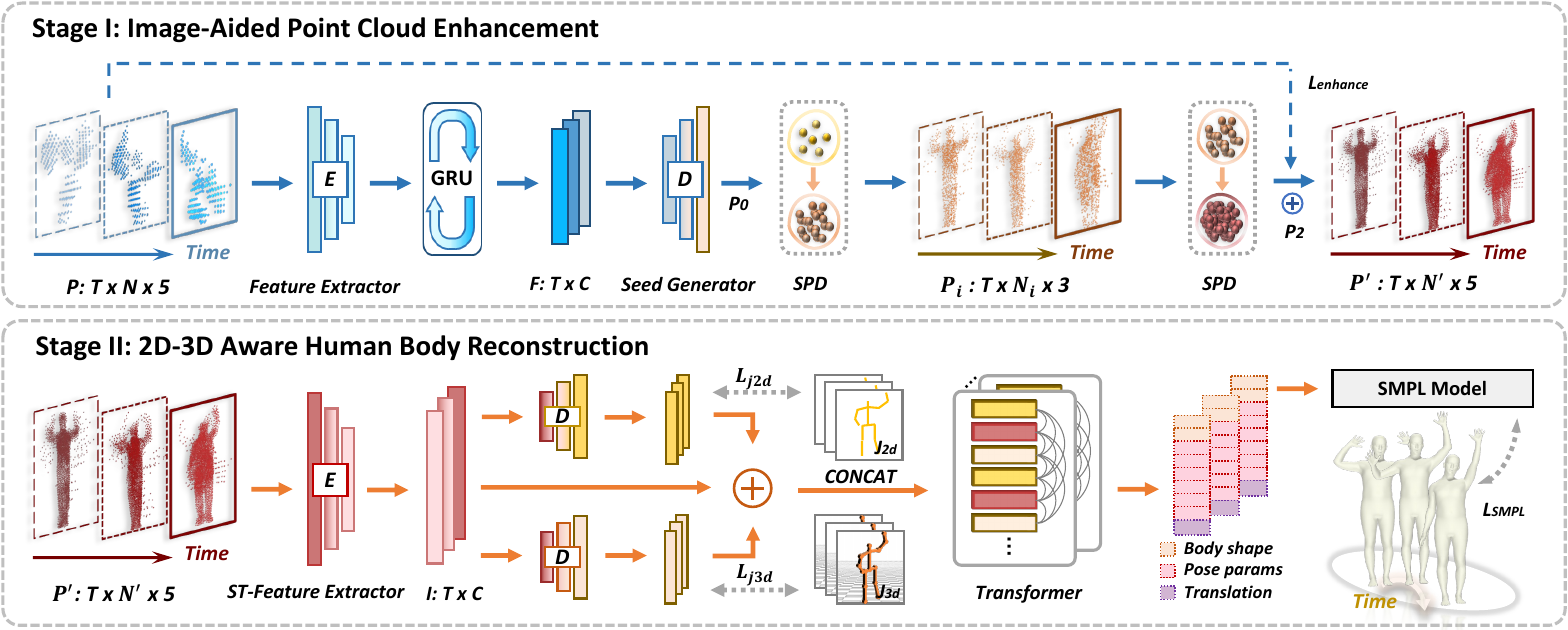} 
    \caption{The framework of our proposed mmDEAR, comprising image-aided point cloud enhancement module and 2D-3D aware human body reconstruction module. These two modules are trained separately, with the trained enhancement module serving as a pre-processing step for the reconstruction module during both the training and inference processes.}
    \label{2}
    \vspace{-5mm}
\end{figure*}

\section{Proposed Method}
In this section, we present our proposed method mmDEAR, a two-stage deep-learning framework for human body reconstruction using mmWave point clouds. As shown in Fig. \ref{2}, our proposed model consists of two modules: 1) Image-aided point cloud enhancement module and 2) 2D-3D aware human body reconstruction module.

In the first stage, the image-aided point cloud enhancement module takes the raw radar point cloud sequence as input and generates a denser and more informative point cloud. This process is supervised by 2D human masks derived from a mature segmentation architecture \cite{ovgseg}, which captures rich posture and shape information.
In the second stage, the 2D-3D aware human body estimation module extracts both 2D and 3D motion features from both the raw and enhanced mmWave point clouds, regressing the 2D and 3D joint positions separately. Then, the 2D and 3D motion information, along with global features, is fused to regress the final SMPL parameters.


\subsection{Problem Formulation}
\label{PS}
Human body reconstruction using mmWave point cloud sequences aims to predict 3D joint locations or generate human parametric models, such as SMPL-X \cite{smplxexpressive}. 
Given a dataset $\mathcal{D} = \{\mathbf{P}, \mathbf{M}, \mathbf{Y}\}$, where $\mathbf{P} \in \mathbb{R}^{T \times N \times 5}$ represents the raw mmWave radar point cloud sequence with time length $T$, $N$ points and 5 channels per frame, the channels contain 3D position, radial velocity, and intensity information. 
$\mathbf{M}\in \mathbb{R}^{T \times N' \times 2}$ is the human mask extracted from the corresponding image frame using OVSeg \cite{ovgseg}, where $N'$ is the sub-sampled point number in the frame. Since the original human mask is in the pixel coordinate system, the depth channel is padded with zeros and transformed to align with the 3D point cloud enhancement module.
In the first stage, raw mmWave point cloud sequence $\mathbf{P}$ serves as the input, with the goal of producing an enhanced point cloud, denoted as $\mathbf{P}'\in \mathbb{R}^{T \times N' \times 5}$.
For the second stage, the enhanced point cloud $\mathbf{P}'$ is utilized to regress the ground-truth SMPL-X parameters $\mathbf{Y}= \{\mathbf{\theta}, \mathbf{\beta}, \mathbf{\gamma}\}$, with the 2D joint positions ${\mathbf{J}}_{2d}$ and 3D joint positions ${\mathbf{J}}_{3d}$ serving as intermediate variables. The pose parameter $\mathbf{\theta}$, shape parameters $\mathbf{\beta}$, and the translation ${\gamma}$ are then fed into an off-the-shelf SMPL model to reconstruct the 3D human joints $\mathbf{J}$ and mesh vertices $\mathbf{V}$. 

\subsection{Image-aided point cloud enhancement module}
Previous mmWave point cloud-based human body reconstruction methods \cite{yang2024mmbat, mmmesh} have shown that the sparsity and noise in raw radar points limit reconstruction accuracy. The sparse mmWave point cloud struggles to fully capture the motion of every body part. Inspired by point cloud completion techniques \cite{xiang2022snowflake, luan2024diffusion}, we leverage human shape information from single-view images as a supervision signal for enhancing mmWave point clouds, which provides more detailed and intuitive shape and structural information. Notably, image data is used only during the training phase, while during inference, only the raw point cloud is used as input, ensuring better privacy protection. To achieve this, we first designed an image-aid mmWave point cloud enhancement module consisting of three sub-modules: 1) spatio-temporal feature aggregation, 2) seed point cloud generation, and 3) cascaded point cloud refinement module.

\par

\textbf{Spatio-temporal feature aggregation:}
As described in Sec. \ref{PS}, this module takes the raw mmWave point cloud $\mathbf{P}\in \mathbb{R}^{T\times N \times 5}$ as input. Since the primary focus of this module is shape information, we first normalize the point cloud and later reverse this process to restore the translation information for the enhanced point cloud.
To extract the global features from the raw point cloud sequence, we use PointNet++ \cite{qi2017pointnet++} as the feature extractor to capture spatial features. Unlike traditional single-frame point cloud methods \cite{xiang2022snowflake, cascaded}, human poses exhibit strong temporal correlations. To leverage this, we apply a bi-directional Gate Recurrent Unit (GRU) to capture the global spatio-temporal feature $\mathbf{F} \in \mathbb{R}^{T\times C}$:
\begin{equation}
\mathbf{F} = \text{GRU}(\text{PointNet++}(\mathbf{P}))
\end{equation}
\par

\textbf{Seed point cloud generation:}
With global spatio-temporal feature $\mathbf{F}$ extracted from raw mmWave point cloud, a seed generator is employed to generate a coarse yet complete seed point cloud $\mathbf{P}_0$. The seed generator refines the global spatio-temporal feature $\mathbf{F}$ into $\mathbf{F}' \in \mathbb{R}^{1 \times C'}$ using an MLP, and concatenates it with $\mathbf{F}$ to form the integrated feature $\mathbf{G} \in \mathbb{R}^{T \times (C + C')}$.
With this integrated feature $\mathbf{G}$, we apply a point-wise splitting operation followed by an MLP-based decoder to regress a candidate point cloud $\mathbf{P}_c$ of size $T\times N_c\times3$. $\mathbf{P}_c$ is then merged with the input point cloud $\mathbf{P}$ and sampled through farthest point sampling (FPS) to generate the seed point cloud $\mathbf{P}_0$.
\par

\textbf{Cascaded point cloud refinement:}
Inspired by \cite{xiang2022snowflake}, we apply two Snowflake Point Deconvolution (SPD) modules \cite{xiang2022snowflake} to the seed point cloud $\mathbf{P}_0$ to produce upsampled point clouds $\mathbf{P}_1$ and $\mathbf{P}_2$, each containing $N_1$ and $N_2$ points per frame, where $\mathbf{P}_i = \text{SPD}(\mathbf{P}_{i-1})$. Finally, to preserve velocity and intensity information from the raw radar points, the upsampled point clouds $\mathbf{P}_2$ are concatenated with raw mmWave points clouds $\mathbf{P}$ and down-sampled to produce the enhanced output $\mathbf{P}' \in \mathbb{R}^{T \times N' \times 5}$.

\par
For the image-aided point cloud enhancement module, we use the 2D human masks from single-view images extracted by the OVSeg \cite{ovgseg} as supervision. The L2 norm Chamfer distance (CD) and partial matching loss \cite{wen2021} are adopted as the loss function, represented by:
\begin{equation}
\footnotesize
\begin{aligned}
    \mathcal{L}_{\text{CD}_2}(\mathcal{P}, \mathcal{M}) &= \sum_{\mathbf{P_i} \in \mathcal{P}} \min_{\mathbf{M_i} \in \mathcal{M}} \|\mathbf{P_i} - \mathbf{M_i}\|_2 +\sum_{\mathbf{M_i} \in \mathcal{M}} \min_{\mathbf{P_i} \in \mathcal{P}} \|\mathbf{M_i} - \mathbf{P_i}\|_2, \\
    \mathcal{L}_{\text{par}}(\mathcal{P}, \mathcal{M}) &= \sum_{\mathbf{P_i} \in \mathcal{P}} \min_{\mathbf{M_i} \in \mathcal{M}} \|\mathbf{P_i} - \mathbf{M_i}\|_2, \\
    \mathcal{L}_{\text{enhance}} &= \mathcal{L}_{CD_2} + \lambda \mathcal{L}_{\text{par}},
\end{aligned}
\end{equation}
where $\mathcal{P}=\{\mathbf{P}_0, \mathbf{P}_1,\mathbf{P}_2\}\,$ denotes the predicted seed points in each step, $\mathcal{M}=\{\mathbf{M}_0, \mathbf{M}_1,\mathbf{M}_2\}\,$ represents the down-sampled ground truth human mask points, with point numbers matching the corresponding predictions. The total loss for the enhancement module is the sum of $\mathcal{L}_{CD_2}$ and $\mathcal{L}_{\text{par}}$, with $\lambda$ typically set to 1.

\subsection{2D-3D aware human body reconstruction module}
With the enhanced point clouds $\mathbf{P}'$ from the previous module, the 2D-3D aware human body reconstruction module aims to regress the SMPL-X parameters $\mathbf{Y}$. Although the enhanced points cloud offers rich shape and pose information, there may be depth ambiguities, especially with overlapping body parts like arms and legs. To address this, it is essential to incorporate the raw mmWave point cloud as input to the model, as it provides the 3D spatial information to help resolve depth ambiguity.
We adopt a multi-stage approach inspired by single-camera body reconstruction techniques \cite{chi2023pose}, where 2D and 3D joint positions are first regressed as an intermediate step before predicting the final 3D SMPL parameters.  

\par
\textbf{Global features extraction with spatio-temporal aggregation:}
The P4Conv architecture \cite{fan2021p4trans}  is leveraged as the backbone of the motion feature extractor, which could effectively extract the information from both raw and enhanced point clouds. Compared with other frame-level point feature extractors, it performs point convolution across both spatial and temporal dimensions, extracting both the representing points $\mathbf{P}_e \in {T\times N_e \times3}$ and the encoded features $\mathbf{F}_e \in{T\times N_e \times C_e}$, where $N_e$ denotes the number of sample points. Furthermore, the temporal embedding is archived by adding the time step to the spatial dimension to formulate  $\mathbf{P}'_e \in {T\times N_e \times 4}$, where 4 strands for  $(x, y, z, t)$. The global feature $\mathbf{I} \in \mathbb{R}^{T \times C_g}$ is then obtained as follows:
\begin{equation}
    \mathbf{I} = \text{MLP}(\mathbf{W} \cdot \mathbf{P}_e' +\mathbf{F}_e)
\end{equation}
where $\mathbf{W}\in C_e \times 4$ is the embedding weight, and the global feature $\mathbf{I}$ is fused with the encoded feature $\mathbf{F}_e$, an MLP layer is adopted to aggregate features in the representing points.
\par

\textbf{2D and 3D joint regression:}
To fully utilize the information in the raw and enhanced point clouds, we regress 2D and 3D joints in parallel. The global feature $\mathbf{I}$ is passed through a linear encoder to obtain the 2D and 3D local features $\mathbf{I}_{2d}$ and $\mathbf{I}_{3d}$, respectively. These local features are then used to regress the 2D joints $\hat{\mathbf{J}}_{2d} \in {T\times N_{J} \times 2}$ and 3D joints $\hat{\mathbf{J}}_{3d} \in {T\times N_{J} \times 3}$ via an MLP decoder. With these intermediate outputs, we apply the L1-norm loss between the predicted joints and the SMPL-X body joints $\mathbf{J}$, represented by:
\begin{equation}
  \mathcal{L}_{\mathbf{J}_{3d}} = \sum \left\lVert \mathbf{J}_{3d} - \hat{\mathbf{J}}_{3d} \right\rVert_1,    \mathcal{L}_{\mathbf{J}_{2d}} = \sum \left\lVert \mathbf{J}_{2d} - \hat{\mathbf{J}}_{2d} \right\rVert_1,
  \label{eq1}
\end{equation}
where $\mathbf{J}_{3d}$ is the SMPL-X body joints $\mathbf{J}$ in 3D space, while $\mathbf{J}_{2d}$ denotes the projection of the raw joints $\mathbf{J}$ onto the x and y axes.

\textbf{Human body reconstruction with 2D-3D aware fusion:}
To dynamically select the informative features from both 2D/3D local and global features, we implement a 2D-3D aware feature fusion module $\Phi$ using a transformer architecture \cite{vaswani2017attention}. The fused feature $\mathbf{I}'$ is computed as:
\begin{equation}
    \mathbf{I}'= \Phi \left( \mathbf{I}_{2d},\mathbf{I}_{3d},\mathbf{I} \right)
\end{equation}
Finally, the fused feature $\mathbf{I}'$ is used to regress the SMPL-X parameters $\hat{\mathbf{Y}} = {\hat{\mathbf{\theta}}, \hat{\mathbf{\beta}}, \hat{\mathbf{\gamma}}}$ through an MLP decoder. These parameters are then passed to the SMPL-X model to compute the body joints $\hat{\mathbf{J}}$ and mesh vertices $\hat{\mathbf{V}}$. The loss function for this stage includes the L1 loss for shape, translation, joints, and mesh vertices, as well as the geodesic loss \cite{geodisic} for the human pose parameter:
\begin{equation}
    \mathcal{L}_{\text{SMPL}} = \mathcal{L}_{\theta} + \mathcal{L}_{\beta} + \mathcal{L}_{\gamma} + \mathcal{L}_{\mathbf{J}} + \mathcal{L}_{\mathcal{V}}
    \label{eq2}
\end{equation}

With the above loss functions introduced in Eq.\ref{eq1} and Eq.\ref{eq2}, we jointly optimize both stages with a total loss defined as:
\begin{equation}
    \mathcal{L}_{\text{all}} = \mathcal{L}_{\mathbf{J}_{3d}}+\mathcal{L}_{\mathbf{J}_{2d}} + \mathcal{L}_{\text{SMPL}}
\end{equation}

\section{EXPERIMENTS}
In this section, we evaluate our proposed method and compare its performance with prevailing SOTA techniques. We also conduct ablation experiments to demonstrate the effectiveness of the individual sub-modules.
\subsection{Datasets and Experiment setup}
\textit{1) Datasets:} We employ two public datasets to evaluate our proposed method, \textbf{mmBody dataset} \cite{chen2022mmbody} and \textbf{MRI dataset} \cite{mri}. The mmBody dataset is the first public dataset offering Motion Capture (MoCap) ground-truth data in the SMPL-X format specifically for mmWave-based human body reconstruction tasks. It includes over 100 motions from 20 subjects, with dense radar point clouds comprising thousands of points per frame. The test set includes scenarios captured in challenging conditions such as darkness, rain and fog, which test the robustness of our method.
\input{table1}
\par
In contrast, the MRI dataset \cite{mri}, which focuses on simple rehabilitation exercises, contains over 160,000 frames from 20 subjects in home monitoring conditions. The ground truth is captured by two Kinect cameras that consists of 17 joint locations. For this dataset, we adapt our final output to the 3D joint locations rather than the SMPL-X model parameters.
\par

\textit{2) Experimental Setup:}
The deep learning architecture is implemented with PyTorch and trained on an NVIDIA GeForce RTX 3090 GPU. We train our model for 100 epochs with an Adam optimizer, and the initial learning rate is set to 1 × $10^{-3}$. Each loss term is normalized to the same scale. The batch size is set to 32 and the time length is 5. The dropout ratio is set to 0.2 for the GRU module in the first module. For the mmBody dataset, we sample 1024 raw radar points per frame, while for the MRI dataset, we sample 256 points per frame. The enhancement module has up-sampling factors of 2, resulting in 2048 enhanced points for the mmBody dataset and 512 enhanced points for the MRI dataset.
\par

\textit{3) Compared methods:}
We have compared our methods with three existing methods: \textbf{P4Trans} \cite{fan2021p4trans}, \textbf{mmMesh} \cite{mmmesh}, and \textbf{mmBaT} \cite{yang2024mmbat}. Specifically, P4Trans is designed for point cloud sequence understanding and 3D action recognition and is also used as the benchmark method on mmBody dataset \cite{chen2022mmbody}. mmMesh and mmBaT are powerful methods designed for mmWave-based human body reconstruction tasks.
\par

\textit{4) Evaluation metrics:} 
\par





Following \cite{mmmesh,chen2022mmbody,yang2024mmbat}, we adopt Mean Per Joint Position Error \textbf{(MPJPE)}, Mean Per Vertice Position Error \textbf{(MPVPE)}, Mean Translation Error \textbf{(MTE)}, Mean Per Joint Rotation Error \textbf{(MPJRE)} to measure prediction error for all body joint, mesh vertices, body translations, and joint rotations, respectively. For the limb performance, we further evaluate Mean Per Upper Limbs Error \textbf{(MPULE)} and Mean Per Lower Limbs Error \textbf{(MPLLE)} stand for the joint error on the arms and legs. Finally, \textbf{Jitter} \cite{jitter, Yi_2022_CVPR,xia2024envposer} indicates the motion smoothness by calculating the mean jerk (time derivative of acceleration) of all body joints. 

\begin{figure*}[ht!]
    \centering
    \includegraphics[width=\linewidth]{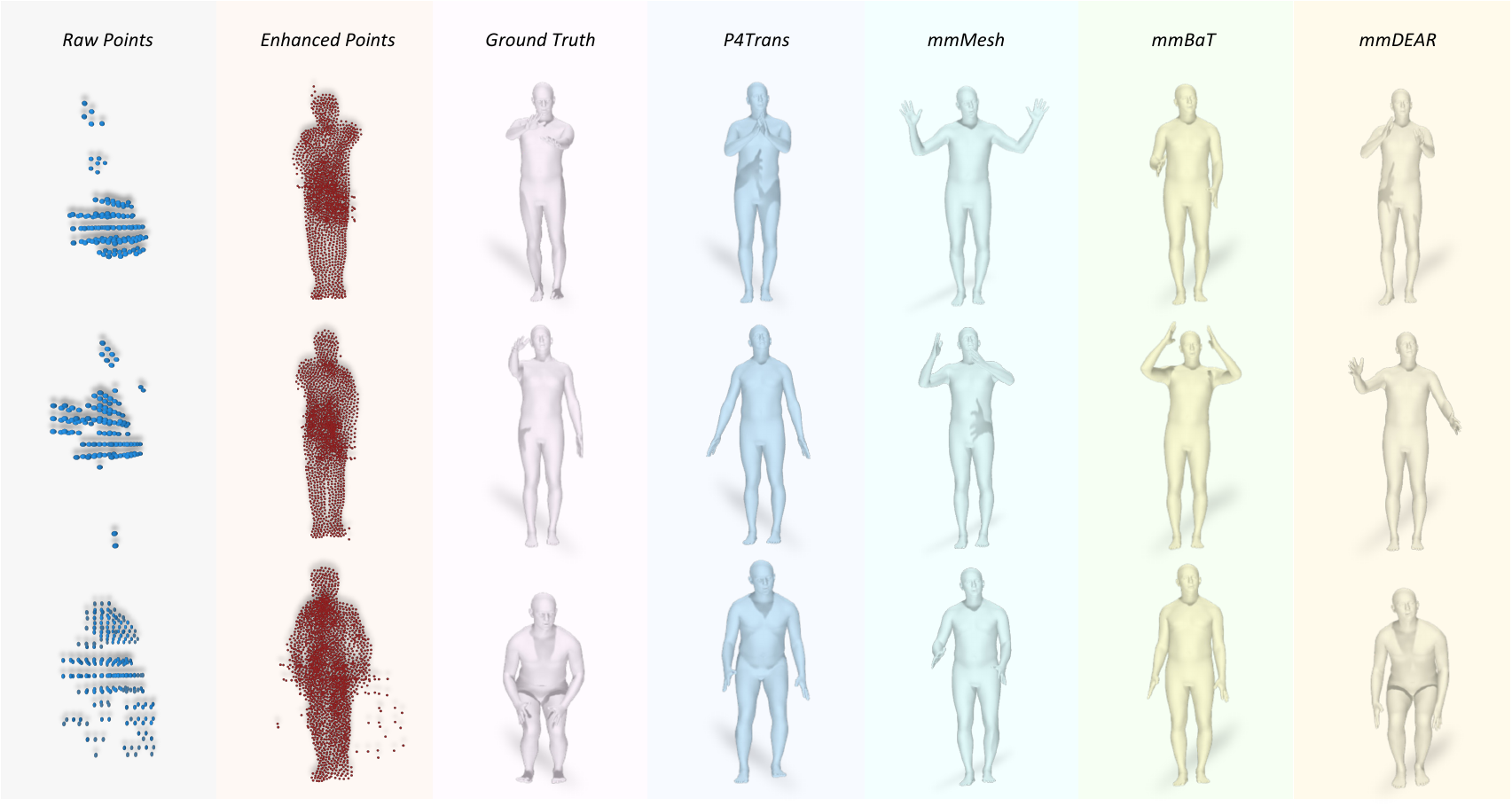} 
    \caption{Qualitative Results for Point Cloud Enhancement and Reconstruction }
    \label{3}
\end{figure*}
\subsection{Main Results}

\textit{1) Quantitative Results:} Table \ref{tab:1} presents the quantitative results of all models evaluated on both datasets. Our proposed method consistently outperforms the existing SOTA methods across nearly all metrics. Specifically, on the mmBody dataset, our method achieves up to 10.58cm and 8.7cm in MPJPE and MPVPE compared to the P4Trans method. We also observe significant improvements in pose error (MPJRE), with up to 9.61$^{\circ}$ better performance compared to mmMesh. The superior performance in MPULE and MPLLE highlights our method's strong ability to capture limb movements. Although the Jitter metric shows only a minor increase, with a 7.1\% difference compared to mmMesh, our method demonstrates substantial advantages of 21.5\% and 20.1\% in MPJPE and MPVPE, respectively.

\par
On the MRI dataset, our method maintains its prominent advantage, showcasing its generalization capability across different mmWave radar datasets with varying radar devices and point cloud densities, especially for the point cloud enhancement module. Due to the simple actions and clear capture environment in this dataset, other methods have relatively acceptable performance, so our method only has limited improvement compared to the mmBody dataset.
\par

\textit{2) Qualitative Results:} Fig. \ref{3} illustrates the point cloud enhancement and body reconstruction results, using samples from the mmBody dataset as a reference. The raw point clouds are quite sparse, with some parts of the human body missing. In contrast, the enhanced point clouds provide a clearer representation of the body shape and limbs. The reconstruction results show that while all methods achieve good regression for the torso, our approach demonstrates a noticeable improvement in capturing the details of the limbs.

\input{table2}
\subsection{Evaluation of the image-aided point cloud enhancement module}

To evaluate the effectiveness and feasibility of the proposed point cloud enhancement module,  we integrated it into the pre-processing pipeline of the compared methods, applying it to both the training and testing phases. The training strategy for these methods was consistent with our approach.  The experiment is conducted on the mmBody dataset and the results are shown in Table \ref{tab2}. The inclusion of our enhancement module led to significant improvements across nearly all metrics for the three compared methods. 
Notably, P4Trans and mmMesh show substantial gains, as they suffer a lot from the sparsity of the raw point cloud. In contrast, mmBaT experiences a slight increase in rotation error due to the introduction of 2D features from the enhanced point cloud. This may have contributed to depth ambiguities, as mmBaT relies solely on 3D joint supervision during the intermediate stages.
\subsection{Ablation Study}
\begin{figure}[ht!]
    \centering
    \includegraphics[width=\linewidth]{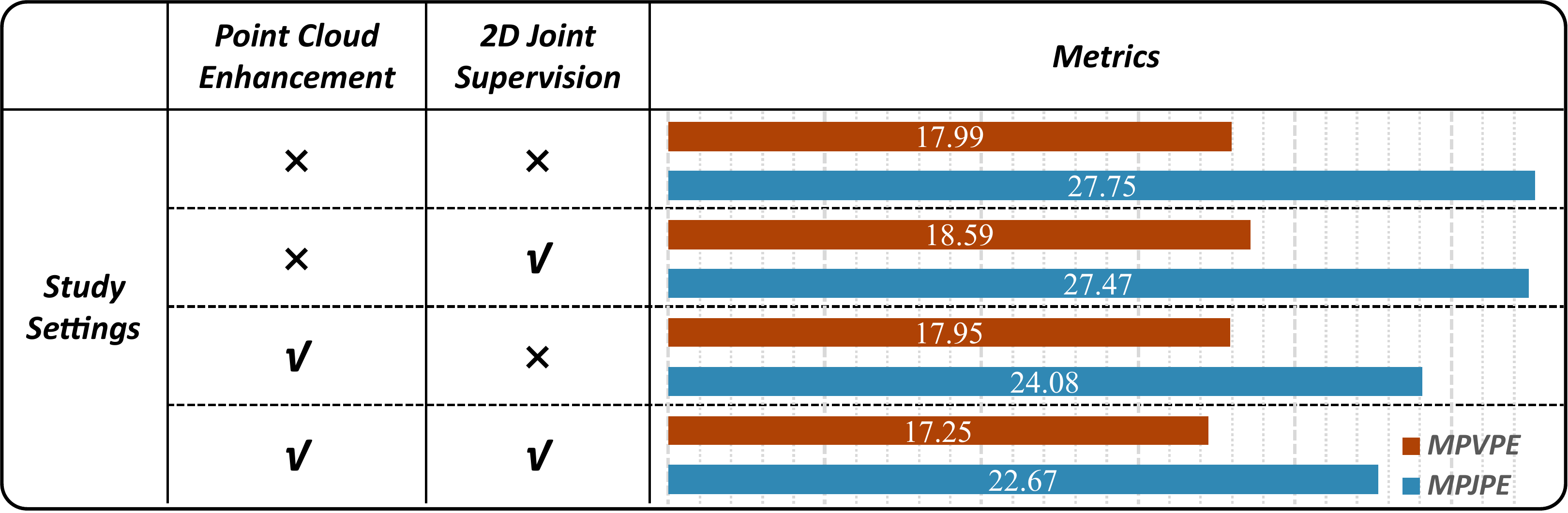} 
    \caption{Ablation Study Results on Point Cloud Enhancement and 2D Joint Supervision}
    \label{fig4}
    \vspace{-3mm}
\end{figure}

\par
The ablation results are illustrated in Fig. \ref{fig4}. Upon removing the enhancement module, we observed a significant degradation in MPJPE and MPVPE, underscoring the critical role of our point cloud enhancement module.
Despite using only the raw point cloud data, our 2D-3D aware body reconstruction module still outperforms vanilla P4Trans and mmMesh, and shows an improvement in MPVPE compared to mmBaT. The slight increase in MPJPE can be attributed to the absence of 2D body posture and shape features, suggesting that 2D joint supervision may not be essential in this context. This is evidenced by further removing the 2D joint supervision, where the results remain similar.
\par
Without the 2D joint supervision, the joint error has a larger increase than the vertice error. This demonstrates the effectiveness of the body reconstruction module and highlights that the design effectively captures detailed 2D information provided by the enhanced point cloud. 

\section{CONCLUSIONS}
In this paper, we introduce mmDEAR, a novel two-stage deep-learning framework designed for human body reconstruction using mmWave point clouds. The first stage features a point cloud enhancement module that employs simple 2D human masks for supervision during training, addressing the inherent challenges of sparsity and inter-frame inconsistency in raw radar point clouds. This module does not necessitate additional image inputs during inference and can be seamlessly integrated into existing methods, enhancing its practical value for real-world applications. In the second stage, our 2D-3D aware human body reconstruction module incorporates 2D and 3D joint regression as intermediate steps, effectively leveraging the enhanced point cloud's detailed shape information alongside the raw point cloud's 3D spatial data.
Our extensive experiments on two public datasets demonstrate significant performance improvements, achieving up to 10.58 cm in MPJPE and 8.7 cm in MPVPE compared to the current state-of-the-art methods.

\bibliographystyle{ieeetr}
\bibliography{root}
\addtolength{\textheight}{-12cm}

\end{document}

%% file: table1.tex
\begin{table*}[ht!]
\centering
\scriptsize
\renewcommand{\arraystretch}{0.8}
\caption{ Comparison with other State-of-the-art methods}
\resizebox{\textwidth}{!}{%
\begin{tabular}{c|c|c|c|c|c|c|c|c}
\toprule
Dataset& Method & \makecell[c]{MPJPE \\($cm$)} & \makecell[c]{MPVPE \\ ($cm$)}  & \makecell[c]{MTE \\ ($cm$)}  & \makecell[c]{MPJRE \\ ($^{\circ}$)} & \makecell[c]{MPULE \\ ($cm$)} & \makecell[c]{MPLLE \\ ($cm$)} & \makecell[c]{Jitter\\ ($km/s^{-3}$)} \\
       
\midrule
\multirow{4}{*}{mmBody \cite{chen2022mmbody}} 
         & P4Trans \cite{fan2021p4trans} &33.25& 25.95 &  16.81& 17.33 & 29.10 & 27.59 & 0.312  \\
         & mmMesh \cite{mmmesh}   & 28.91 &21.60&15.46& 22.08 & 24.84 & 22.17 &  \textbf{0.118}  \\
         & mmBaT \cite{yang2024mmbat}   & 25.10 & 19.91 & 14.54 & 12.83 & 23.07 & 19.85 &  0.135  \\
         & mmDEAR    & \textbf{22.67}  & \textbf{17.25} & \textbf{11.99} & \textbf{12.47} & \textbf{19.03} & \textbf{11.98} &  0.127  \\
\midrule
\multirow{4}{*}{MRI \cite{mri}} 
          & P4Trans \cite{fan2021p4trans} & 12.98 & / & 8.46 & 10.82 & 13.47 & 10.99 & 0.175 \\
         & mmMesh \cite{mmmesh}    & 12.46 & / & 8.41 & 11.07 & 14.33 & 10.04 & 0.134  \\
         & mmBaT \cite{yang2024mmbat}    & 9.73 &  /& 7.76 & \textbf{9.03} & 12.47 & 8.99 & 0.149 \\
         & mmDEAR    &\textbf{9.42}  & / & \textbf{7.54} & 9.84 & \textbf{11.69} & \textbf{8.33} &  \textbf{0.125} \\
\bottomrule
\end{tabular}
}
\label{tab1}

\begin{tablenotes}
\normalsize 
\item Note: MPVPE is not compared in the MRI dataset since mesh vertices are not provided. 
\end{tablenotes}
\vspace{-5mm}

\label{tab:1}
\end{table*}

%% file: table2.tex
\begin{table}[htb]
\centering
\footnotesize
\resizebox{\linewidth}{!}{
\begin{threeparttable}[b]

\caption{Performance of the Compared Methods with the Enhanced Point Clouds}
\renewcommand{\arraystretch}{1.2}
\begin{tabular}{l|c|c|c|c}
\toprule
\makecell[l]{Evaluation \\ Metrics}& \makecell{P4Trans \\ \cite{fan2021p4trans}} & \makecell{mmMesh \\ \cite{mmmesh}} & \makecell{mmBaT \\ \cite{yang2024mmbat}} & mmDEAR \\
\midrule
MPJPE ($cm$)       & 29.88  & 25.61 & 24.17 & \textbf{22.67} \\
MPVPE ($cm$)       & 24.08  & 19.08 & 19.36 & \textbf{17.25} \\
MTE ($cm$)         & 17.51  & 14.24 & 15.47 & \textbf{11.99} \\
MPJRE ($^\circ$)   & 13.41  & 13.59 & 12.79 & \textbf{12.47} \\
MPULE ($cm$)       & 25.50  & 20.89 & 25.17 & \textbf{19.03} \\
MPLLE ($cm$)       & 20.43  & 14.16 & 25.17 & \textbf{11.98} \\
Jitter ($km/s^{-3}$)& 0.336  & 0.136 & 0.131 & \textbf{0.127} \\
\bottomrule
\end{tabular}
\vspace{-3mm}
\label{tab2}
\end{threeparttable}}
\end{table}
